\documentclass[11pt]{article}

\usepackage[final]{acl}

\usepackage{times}
\usepackage{latexsym}

\usepackage[T1]{fontenc}

\usepackage[utf8]{inputenc}

\usepackage{microtype}

\usepackage{inconsolata}

\usepackage{graphicx}

\usepackage{amsmath,booktabs,caption,cleveref,enumitem,multirow,subcaption,xspace}
\usepackage[mode=match]{siunitx}

\setlist{noitemsep,nolistsep}
\DeclareSIUnit[quantity-product = ]\percent{\char`\%}  
\robustify\bfseries  

\title{On the Interplay between Human Label Variation and Model Fairness}

\author{
  \textbf{Kemal Kurniawan}\textsuperscript{1}\hspace{4mm}
  \textbf{Meladel Mistica}\textsuperscript{1}\hspace{4mm}
  \textbf{Timothy Baldwin}\textsuperscript{2,1}\hspace{4mm}
  \textbf{Jey Han Lau}\textsuperscript{1}
  \\
  \textsuperscript{1}School of Computing and Information Systems, University of
  Melbourne \\
  \textsuperscript{2}Mohamed bin Zayed University of Artificial Intelligence
  \\
  \texttt{\{kurniawan.k,misticam,laujh\}@unimelb.edu.au}\hspace{8mm}\texttt{tb@ldwin.net}
}

\newcommand{\methname}[1]{\textsc{#1}}  
\renewcommand{\subset}{subset\xspace}  
\newcommand{\subsets}{subsets\xspace}  
\newcommand{\Subset}{Subset\xspace}  

\begin{document}
\maketitle
\begin{abstract}
The impact of human label variation (HLV) on model fairness is an unexplored topic. This paper examines the interplay by comparing training on majority-vote
labels with a range of HLV methods. Our experiments show that without explicit
debiasing, HLV training methods have a positive impact on
fairness under certain configurations.\footnote{Code and supplementary materials:
  \url{https://github.com/kmkurn/hlv-fairness-interplay}}
\end{abstract}

\section{Introduction}

Human label variation~\citep[HLV;][]{plank2022} has been shown to improve performance through better generalisation~\citep{peterson2019,uma2021,kurniawan2025}. However, the impact
of HLV on model fairness remains unexplored. 
One might conjecture that HLV would improve fairness because it
preserves minority views and avoids collapsing them into a single point like
majority voting.
However, it is well established that there is a trade-off 
between performance and
fairness~\citep{han2022a,shen2022}, so one might also conjecture that HLV would similarly negatively impact fairness (since HLV improves performance).
We explore this interplay between HLV and model fairness in this paper, in what we believe to be the first systematic analysis of this interaction. We compare
training on majority-vote labels with four HLV training methods
from prior work~\citep{kurniawan2025}, and evaluate on two tasks:
offensiveness classification with SBIC~\citep{sap2020} and legal area
classification with TAG, an in-house legal dataset, developed with practising lawyers.

We find that across SBIC and TAG, training with HLV methods improves overall
performance without harming fairness. Over TAG, training with HLV even \emph{improves}
fairness.
Since these findings are only from a single set of class and group
weights~(\textit{configuration} hereafter), we examine many possible
configurations to draw more robust conclusions. For both datasets, HLV
does not reduce model fairness for most configurations, and this trend is
clearer when the configuration prioritises all groups/classes equally.
Interestingly, HLV can improve model fairness for a substantial
number of configurations.

We then investigate why HLV improves fairness by introducing temperature
scaling to an HLV training method to control the weighting of minority annotations. Our TAG experiments suggest fairness gains do indeed come from minority annotations.

Our findings suggest that training with HLV improves performance
without compromising fairness. Additionally,
our findings underscore the effect of class and group weightings in
measuring fairness in the HLV context. Therefore, it is important to understand what fairness means for each application before incorporating HLV.

\section{Related Work}

HLV~\citep{plank2022} challenges the standard
assumption that an instance has one ground truth, in line with data perspectivism~\citep{cabitza2023} which advocates for the
consideration of multiple perspectives in data
annotation.
Previous work on this topic includes model
generalisation~\citep{uma2021,uma2021b,leonardelli2023}, evaluation
metrics~\citep{uma2021,rizzi2024,kurniawan2025}, model
calibration~\citep{baan2022}, and separating annotation signal from
noise~\citep{weber-genzel2024a,ivey2025}. HLV has also been shown to improve
model generalisation for computer vision tasks~\citep{peterson2019}. Our work
contributes to this growing discipline by examining the interplay
between HLV and fairness.

Previous work covers the interplay between model fairness and other
aspects of machine learning models. \citet{hessenthaler2022a} studied how model
fairness and environmental costs affect each other. \citet{zhao2023a,brandl2024}
explored the interaction between model fairness and explainability, as did
\citet{ferry2025} but with privacy considerations. In contrast, we
study how model fairness is impacted by HLV, which to our knowledge, is the first examination of this interplay.

\section{Method}

\subsection{Datasets}\label{sec:datasets}

\paragraph{SBIC}

Each instance in the Social Bias Inference Corpus~\citep{sap2020} is an English social media post annotated with an
offensiveness class and a list of target minority groups.\footnote{Offensiveness classes = Yes/No/Maybe; example target groups = black folks, asian folks, etc.} Target groups are
divided into 7 broad categories\footnote{Body, culture, disabled, gender,
  race, social, victim.} with each post annotated by 3 crowd annotators.
We compute its offensiveness class distribution and union of the broad
categories over these 3 annotations.

\paragraph{TAG}

A private dataset developed together with Justice Connect,\footnote{\url{https://justiceconnect.org.au}} a legal non-profit that helps connect legal help-seekers to pro-bono lawyers. The dataset is drawn from an intake triage process, where each instance is an English description of a legal problem by a help-seeker and
annotated with one or more areas of law (e.g.,\ \textit{Criminal law},
\textit{Not a legal issue}) by practising lawyers, rendering the task
multi-label.
In addition, each instance is associated with one or more
self-reported help-seeker demographic identities/cohorts (e.g.,\ seniors, low income earners).\footnote{More details~(incl.\ splits) are given in \Cref{sec:TAG-details}.}
Justice Connect provides a weighting for
the classes (i.e.\ legal areas) and groups (i.e.\ cohorts) reflecting their priorities in the intake process.
We use this weighting for the fairness evaluation
in \Cref{sec:eval}. The average inter-annotator agreement\footnote{Computed on
  \qty{10}{\percent} of the data due to computational costs.}~over
the legal areas is Krippendorf's $\alpha$ = \num{0.454}, which is modest. 
Manual analyses and discussions with the organisation confirm that the annotation disagreements are often valid, suggesting they are genuine variations rather than noise.

\subsection{Evaluation}\label{sec:eval}

We focus on a \emph{group-level parity} definition of fairness in this study due to its
prevalence in prior work~\citep{shen2022,han2023}. To evaluate this, we first compute a
fairness score $s_{kg}$ (defined later) for each class $k$ and group $g$. Next,
following \citet{han2023}, we aggregate $s_{kg}$ over groups then classes using
weighted generalised mean to obtain the overall fairness score. The group-wise
aggregation is defined as
\begin{equation}
  \bar{s}_k=\left(\sum_gw_gs_{kg}^p \right)^{\frac{1}{p}}\label{eqn:grp-agg}
\end{equation}
where $p\neq0$ is a scalar exponent controlling the contribution of
smaller-valued $s_{kg}$ and $w_g$ is a
non-negative importance weights for group $g$ satisfying $\sum_gw_g=1$. The
class-wise aggregation is performed analogously over $\bar{s}_k$ using separate
class-specific weights and exponent.

To define the groups, we use the broad target
categories for SBIC, and the defined cohorts for TAG. Therefore, for SBIC, the
group information is associated with the \emph{target} of the social media post, while the group information for TAG is associated with the
\emph{help-seeker} who \emph{produced} the instance.

In both datasets, each instance can belong to zero
or more groups. Therefore, we partition the instances into two \subsets for
each group based on in- or out-of-group membership, respectively.\footnote{See \Cref{sec:subset-size-stats} for
  \subset size statistics.} We compute $s_{kg}$ by first evaluating performance
for class $k$ on both \subsets then taking the ratio between the worse- and the better-performing \subsets,
following \citet{yeh2024}.\footnote{See \Cref{sec:fair-eval} for a more formal definition.}

Computing $s_{kg}$ requires a performance metric for each class that works in the HLV context~(i.e., soft ground truths).
Thus, we use soft F\textsubscript{1}, the only metric satisfying the requirement~\citep{kurniawan2025}.
For a given class $k$, it measures the amount of overlap between the
ground-truth and predicted probabilities of class $k$ relative to their total.\footnote{See \Cref{sec:perf} for details.}

\paragraph{Aggregation configurations}
For TAG results in \Cref{sec:results}, the class and
group aggregation weights $w_g$ are provided by the legal organisation who co-developed the dataset (\Cref{sec:datasets}). For SBIC, as we have no a priori preference we weight each class and group
equally. For both
datasets, we set $p=1$ for both aggregations, resulting in standard weighted average.
In \Cref{sec:mult-weight}, we experiment with multiple randomly-sampled
configurations for more robust conclusions.\footnote{Details are given in
\Cref{sec:conf-samp}.}

\paragraph{Overall performance}
To assess performance in the HLV context, we use soft \emph{micro}
F\textsubscript{1}, a natural extension of class-wise soft
F\textsubscript{1} which correlates well with human
judgement~\citep{kurniawan2025}.\footnote{See \Cref{sec:perf} for details.}

\subsection{Approaches}

We experiment with two pretrained models: base RoBERTa~\citep[$\approx$100M parameters]{liu2019h} and
Nov'24 7B OLMo~2~\citep{walsh2025}. We replace the output layer of each model
with the appropriate classification layer\footnote{A linear layer then
  either a softmax~(SBIC) or sigmoid~(TAG) activation functions.} and
finetune on the train set.

For HLV training methods, we use the four most successful methods
identified by \citet{kurniawan2025}: repeated labelling~(\methname{ReL}), soft
labelling~(\methname{SL}), minimising Jensen-Shannon
divergence~(\methname{JSD}), and maximising soft micro F\textsubscript{1}
score~(\methname{SmF1}).\footnote{See \Cref{sec:hlv-train} for more details.} For non-HLV baselines, we finetune them on majority-vote classes (\methname{MV}).\footnote{Further training details are reported in \Cref{sec:trn-details}.}

\section{Results}

\subsection{Single Fairness Configuration}\label{sec:results}

We report the overall performance and fairness of each method in
\Cref{tbl:results}. For SBIC, the table shows that HLV training methods
consistently outperform \methname{MV} for overall performance across the
two models. These improvements are statistically\footnote{We report effect sizes
as a measure of \emph{practical} significance in \Cref{sec:cies}.} significant~(with bootstrap tests) for 6 out
of 8 model--method pairs.\footnote{Exceptions: \methname{SL} and \methname{JSD}
  with RoBERTa} As for fairness, the HLV training methods can
improve or reduce fairness, but the differences are not statistically
significant for 7 out of 8 model--method pairs.\footnote{Exception:
  \methname{SmF1} with RoBERTa}

Looking at the TAG results, for both overall
performance and fairness, the HLV training methods consistently outperform
\methname{MV} across both models, with statistically
significant improvements for 6 out of 8 model--method pairs.\footnote{Exception:
  \methname{SmF1}} Taken together, the results suggest that HLV training
improves overall performance without sacrificing fairness, and in some cases (e.g.,\ TAG), can even improve fairness.

\paragraph{Discussion}

Unsuprisingly, the different HLV methods attain different levels of fairness.
However, these variations are consistent with their overall performance. For example,
\methname{ReL}, a strong method performance-wise, is also among the strongest
in terms of fairness, while \methname{SmF1} is among the weakest. That
\methname{SmF1} attains poor performance is not surprising because
\citet{kurniawan2025} showed that the method can result in degenerate
predictions where all probability mass is concentrated on the majority class.
Looking into class- and group-wise performance for each method, we find that all
noticeable differences seem to align well with overall performance trends.

\begin{table}\small
  \centering
  \sisetup{detect-all=true, table-format=2.1, table-space-text-post={$^*$}}
  \begin{tabular}{@{}llSSSS@{}}
  \toprule
  \multirow[c]{2.5}{*}{Model} & \multirow[c]{2.5}{*}{Method} & \multicolumn{2}{c}{SBIC} & \multicolumn{2}{c}{TAG} \\
  \cmidrule(lr){3-4} \cmidrule(l){5-6}
  {}                          & {}              & {Perf}             & {Fair}   & {Perf}             & {Fair}             \\
  \midrule
  \multirow[c]{5}{*}{OLMo}    & \methname{MV}   & 75.3               & 45.7     & 64.4               & 62.3               \\
  \cmidrule(l){2-6}
                              & \methname{ReL}  & \bfseries 78.9$^*$ & 46.0     & \bfseries 70.4$^*$ & \bfseries 72.8$^*$ \\
                              & \methname{SL}   & \bfseries 78.3$^*$ & 46.3     & \bfseries 69.0$^*$ & \bfseries 72.2$^*$ \\
                              & \methname{JSD}  & \bfseries 78.6$^*$ & 44.2     & \bfseries 66.3$^*$ & \bfseries 68.9$^*$ \\
                              & \methname{SmF1} & \bfseries 79.6$^*$ & 42.6     & 64.8               & 64.7               \\
  \midrule
  \multirow[c]{5}{*}{RoBERTa} & \methname{MV}   & 77.8               & 42.9     & 64.5               & 67.3               \\
  \cmidrule(l){2-6}
                              & \methname{ReL}  & \bfseries 78.7$^*$ & 41.9     & \bfseries 71.5$^*$ & \bfseries 72.8$^*$ \\
                              & \methname{SL}   & 78.0               & 48.5     & \bfseries 70.6$^*$ & \bfseries 72.3$^*$ \\
                              & \methname{JSD}  & 78.1               & 43.1     & \bfseries 70.7$^*$ & \bfseries 71.1$^*$ \\
                              & \methname{SmF1} & \bfseries 79.1$^*$ & 39.2$^*$ & 66.9               & 70.4               \\
  \bottomrule
  \end{tabular}
  \caption{Mean performance~(Perf) and fairness~(Fair) of
    the \methname{MV} baseline and HLV training methods over 3 runs.
    Asterisks~($^*$) indicate statistical
    significance~($p<0.05$) against \methname{MV} with a two-tailed bootstrap
    test~\citep{mackinnon2009}. Significant gains over \methname{MV} are
    boldfaced.}\label{tbl:results}
\end{table}

\subsection{Multiple Fairness Configurations}\label{sec:mult-weight}

The results in \Cref{sec:results} were obtained with one configuration, i.e.\ fixed values for $w_g$ and $p$ in \Cref{eqn:grp-agg} for both group- and
class-wise aggregation.
Here, we test multiple randomly-sampled configurations to draw more
robust conclusions.\footnote{See \Cref{sec:conf-samp} for more details.} This experimental
setup is especially important for SBIC: the single configuration used in
\Cref{sec:results} is divorced from any intended application of the model.

For each randomly-sampled configuration,\footnote{Note that performance is
  constant regardless of configurations as they configure only the fairness score computation.} we perform a
two-tailed bootstrap test 
across 3 runs to
check if the mean fairness score difference between each HLV training method and
\methname{MV} is statistically significant~($p<0.05$). If so, we record if 
the mean fairness score of \methname{MV} is higher~(i.e.\ fairer). Otherwise, we deem the HLV training method as fair as
\methname{MV}. Note that overall performance is unaffected by fairness
configurations. Therefore, we compare only the fairness scores.

\begin{figure}[t]
  \centering
  \includegraphics[width=0.95\columnwidth]{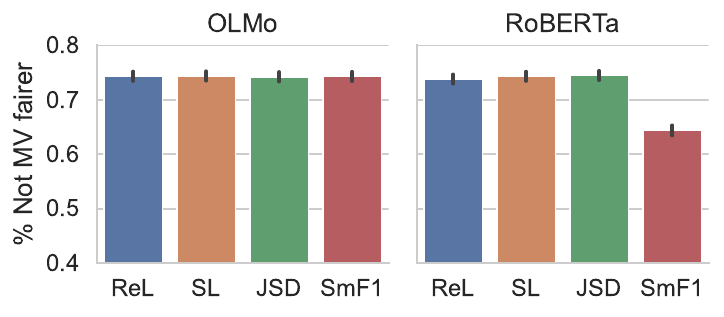}
  \caption{Fraction of randomly sampled configurations where \methname{MV} is
    not significantly fairer than each HLV training method on SBIC.
  }\label{fig:fair-conf-anal-sbic}
\end{figure}

We report the fraction of configurations where \methname{MV} is
found \emph{not} to be fairer than each HLV method on SBIC in
\Cref{fig:fair-conf-anal-sbic}. The result on TAG follows the same trend, which
we show in \Cref{sec:mult-weight-more}.

\Cref{fig:fair-conf-anal-sbic} shows for most model--method pairs, training
with HLV does not hurt model fairness in over \qty{70}{\percent} of
configurations. An exception is \methname{SmF1} with RoBERTa, but the proportion is
still over \qty{60}{\percent}. Remarkably, we find that training with HLV
improves fairness for up to \qty{25}{\percent} of configurations~(see
\Cref{sec:mult-weight-more}). Overall, the results suggest that training with
HLV generally does not hurt fairness, further supporting our findings in
\Cref{sec:results}.

That said, \Cref{fig:fair-conf-anal-sbic} also shows a substantial number of
configurations where \methname{MV} is fairer than training with HLV. This
finding highlights the need to determine fairness
configurations specific to the application before incorporating HLV.

\section{Analysis}

\subsection{Fairness Configurations}\label{sec:fair-conf-anal}

\begin{figure}[t]
  \centering
  \includegraphics[width=0.95\columnwidth]{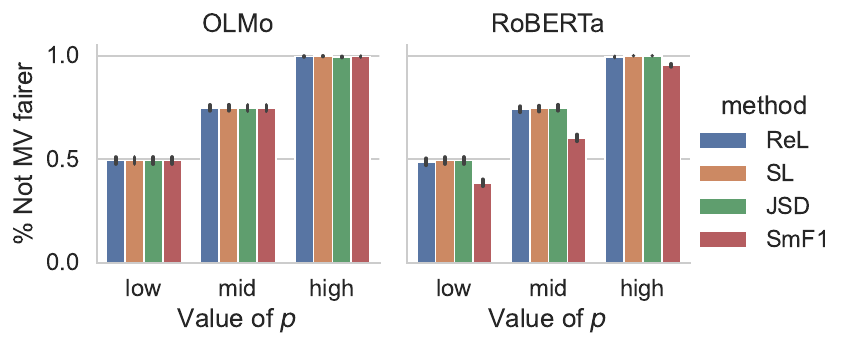}
  \caption{Fraction of randomly sampled configurations where \methname{MV} is
    not significantly fairer than each HLV training method on SBIC for various
    levels of exponent $p$ in the group-wise aggregation.
  }\label{fig:fair-conf-anal-p-grp-sbic}
\end{figure}

This section investigates the role $p$ in \Cref{eqn:grp-agg}
for both group- and class-wise aggregation. We
categorise the value of $p$ into 3 levels: low~($<-5$),
mid~(\numrange{-5}{5}), and high~($>5$). In
\Cref{fig:fair-conf-anal-p-grp-sbic}, we plot the fraction of configurations
where training with HLV does not decrease fairness for each level of $p$ in the
group-wise aggregation on SBIC. \Cref{fig:fair-conf-anal-p-grp-sbic} shows that when $p$ is low, the fraction is about \qty{50}{\percent}, lower than when $p$ is mid-ranging. In contrast, the fraction increases to almost \qty{100}{\percent} when
$p$ is high. A nearly identical trend is observed for the class-wise
aggregation~(see \Cref{sec:fair-conf-anal-more}). TAG results are similar to SBIC~(see \Cref{sec:fair-conf-anal-more}).

Scalar $p$ controls the contribution of
smaller-valued $s_{kg}$ in group-wise (\Cref{eqn:grp-agg}) and class-wise aggregation. Smaller/larger $p$ means prioritising lower-scoring groups or
classes more/less.
Therefore, the results suggest that when one cares
about all classes or groups similarly regardless of their scores~(mid), training with HLV does
not reduce fairness.
This setting corresponds to most existing fairness
evaluation~\citep{han2023}.
However, when one prioritises the lowest-scoring classes or groups in
fairness~(low, similar to
\citet{rawls2001}), training with HLV is comparable to
\methname{MV}.\footnote{The results also suggest that training with HLV
  generally never harms fairness when one focuses more on the highest-scoring classes
  or groups~(high). However, this overestimates model fairness and seems unused in
  practice.}

\subsection{Minority Annotations}\label{sec:minor-anns}

Here, we investigate factors that make HLV beneficial for model
fairness. We focus on the TAG dataset on which HLV generally outperforms
\methname{MV} in both performance and fairness as reported in
\Cref{sec:results}. We hypothesise that the fairness improvements come from the
minority annotations.

To test this, we introduce temperature scaling to \methname{SL}.
Each class probability is proportional to the number of
annotators that select the class raised to the power of $\frac{1}{\tau}$ where
$\tau>0$ is the temperature. As $\tau\to0$, the most-voted class will have
probability close to 1, resembling \methname{MV}. In contrast, the class
distribution will reduce to uniform as $\tau\to\infty$, thereby weighting up the
minority annotations.\footnote{In practice, distributions will be
  uniform over all most-voted classes as
  $\tau\to0$ and over selected classes as
  $\tau\to\infty$.
} Standard \methname{SL}
corresponds to $\tau=1$.

\begin{figure}[t]
  \centering
  \includegraphics[width=0.95\columnwidth]{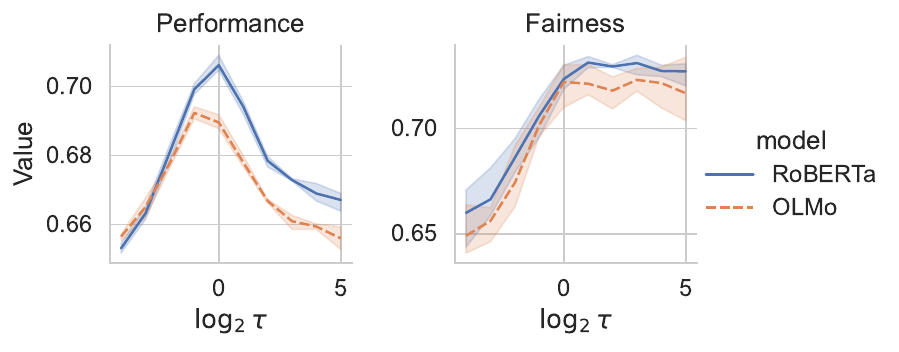}
  \caption{Impact of minority annotations~($\tau$) on fairness~(right) on TAG.
    Larger $\tau$ means minority annotations are weighted more. Impact on
    performance~(left) is shown for completeness.
  }\label{fig:results-alpha}
\end{figure}

We report how fairness changes as $\tau$ grows in
\Cref{fig:results-alpha}. The
figure shows that when $\tau$ is small, fairness is low. As $\tau$ increases,
fairness improves and stabilises after peaking at $\tau=1$. These findings
suggest minority annotations contain useful signals for
fairness. We offer a possible explanation for this in
\Cref{sec:expl-minor-anns-TAG} and leave the full investigation on this
direction for future work.

Unsurprisingly, the figure also shows that performance
peaks at $\tau=1$. This is because the scaling is only applied to the train set, i.e.
test ground truths are computed normally~($\tau=1$).

\section{Conclusions}

While HLV has been shown to improve generalisation, this is the first work to study the interplay between HLV and
fairness. We first showed empirically that training with HLV does not harm model
fairness, and indeed, it can improve fairness. We then showed that this observation is robust to various fairness configurations.
However, when the lowest-scoring groups or classes in fairness were
prioritised, training with HLV provided no fairness benefits compared to
training on majority-vote classes. Lastly, we confirmed that the fairness
improvements with HLV training come from the minority annotations.

Our findings
highlight the performance and fairness benefits of training with HLV under
certain configurations. Therefore, we
recommend NLP practitioners to: (1)~consider what fairness means in their
application, (2)~define it in terms of the
fairness evaluation configuration described in \Cref{sec:eval}, (3)~incorporate HLV
if the configuration prioritises all groups/classes roughly equally, and
(4)~evaluate fairness with the configuration to confirm that HLV does not harm
fairness compared to majority voting.

\section*{Limitations}

Because the TAG dataset contains real-world confidential legal help requests, we cannot distribute
it as it used in the study. However, we still include it because of its importance
to our analysis in \Cref{sec:minor-anns}, and we believe that the insight
obtained from the dataset in the study is still useful. Although we cannot
distribute the dataset, we can distribute ground-truth and predicted judgement
distributions of the dev and the test
sets.\footnote{\url{https://github.com/kmkurn/hlv-fairness-interplay}} This will
allow other researchers to replicate our reported numbers and build on our work
to some extent.

Prior work has evaluated fairness using synthetic, targeted check
lists~\citep{manerba2021}. Therefore, a reasonable solution may be to follow a
similar approach to address the data privacy issue. We agree in theory that it
would allow us to not only measure fairness at a high level but
also understand in what cases the models break down. However, our goal is to
understand fairness and its relationship with HLV rather than address specific impacts of model bias. Furthermore, we
foresee a few complications: (1) it is not always immediately clear what the
biases might be as we need to know them in order to create the synthetic check
lists); (2) in our case, the language of the TAG data is non-standard English,
and so it is difficult to mimic that genre; and (3) we will need to distribute our
TAG-trained models, which is restricted because of IP constraints and possible risks of
leaking the sensitive TAG training data through the model. Therefore, this work
does not take this approach.

We evaluate using soft evaluation metrics in all experiments. One could argue
that this setup favours HLV training, so the comparisons with \methname{MV} are
unfair. However, we note that previous work has found that
\methname{MV} outperforms many HLV training methods in this
setup~\citep{kurniawan2025}, suggesting that the argument does not hold in practice.

Another limitation of our study is the exclusive use of English datasets. This
is because of the scarcity of datasets that have both disaggregated annotations
for HLV and instance attributes (e.g., cohort information) which are appropriate for fairness evaluation.
However, we expect that our findings would generalise to non-English languages,
as our experimental design is language-independent. We encourage future work to
develop such datasets for non-English languages to test this empirically.

\section*{Acknowledgments}

This research is supported by the Australian Research Council Linkage Project
LP210200917\footnote{\url{https://dataportal.arc.gov.au/NCGP/Web/Grant/Grant/LP210200917}}
and funded by the Australian Government. This research is done in collaboration
with Justice Connect, an Australian public benevolent institution.\footnote{As
  defined by the Australian
  government:~\url{https://www.acnc.gov.au/charity/charities/4a24f21a-38af-e811-a95e-000d3ad24c60/profile}}
We thank Kate Fazio, Tom O'Doherty, and Rose Hyland from Justice Connect for
their support throughout the project. This research is supported by The
University of Melbourne’s Research Computing Services and the Petascale Campus
Initiative.

\bibliography{fair}

\appendix

\section{TAG Dataset Details}\label{sec:TAG-details}

The TAG dataset has a total of 11K instances split randomly 8:1:1 for train,
dev, and test sets. Each instance is a legal problem description, e.g.,
\textit{my landlord evicted me w/o notice}, annotated by an
average of 5.5 lawyers with one or more legal areas out of 33 options, e.g.,
\textit{Criminal law}, \textit{Elder law}, \textit{Not a legal issue}. Each
instance is also associated with one or more cohorts out of 6
choices, e.g., seniors, low income earners. Sensitive identifying information
had been anonymised by the owner organisation prior to the dataset use in our work.

\section{\Subset Size Statistics}\label{sec:subset-size-stats}

\Cref{tbl:subset-sizes} reports the number of instances in the in- and
out-of-group \subsets for each group in the datasets. For TAG, the help-seekers were
allowed to select ``Prefer not to say'' when self-reporting. For fairness evaluation, we include only those that did not select
that option, i.e.\ their identities were given explicitly.

\begin{table}\small
  \centering
  \begin{subtable}{\columnwidth}
    \centering
    \begin{tabular}{@{}lrr@{}}
      \toprule
      \multirow[c]{2.5}{*}{Group} & \multicolumn{2}{c}{Subset} \\
      \cmidrule(l){2-3}
               & {In-group} & {Out-of-group} \\
      \midrule
      Body     & 58         & 4640           \\
      Culture  & 495        & 4203           \\
      Disabled & 112        & 4586           \\
      Gender   & 503        & 4195           \\
      Race     & 819        & 3879           \\
      Social   & 104        & 4594           \\
      Victim   & 215        & 4483           \\
      \bottomrule
    \end{tabular}
    \caption{SBIC}
  \end{subtable}

  \vspace{4mm}

  \begin{subtable}{\columnwidth}
    \centering
    \begin{tabular}{@{}lrr@{}}
      \toprule
      \multirow[c]{2.5}{*}{Group} & \multicolumn{2}{c}{Subset} \\
      \cmidrule(l){2-3}
          & {In-group} & {Out-of-group} \\
      \midrule
      ATS & 35         & 317            \\
      HOM & 154        & 64             \\
      LGB & 29         & 8              \\
      LOW & 135        & 148            \\
      PUB & 19         & 109            \\
      SEN & 164        & 90             \\
      \bottomrule
    \end{tabular}
    \caption{TAG (ATS = indigenous, HOM = experiencing or at risk of
      homelessness, LGB = LGBTQ+ individuals, LOW = low-income earners, PUB =
      public housing dwellers, SEN = seniors)}
  \end{subtable}
  \caption{Number of instances in the in- and out-of-group \subsets defined by
    each group in SBIC~(top) and TAG~(bottom).}\label{tbl:subset-sizes}
\end{table}

\section{Evaluation Details}

\subsection{Definition of $s_{kg}$}\label{sec:fair-eval}

If $F_{kg}^1$ and $F_{kg}^0$ denote the performance for class $k$ on the
in- and out-of-group \subsets defined by group $g$ respectively, then the
fairness score $s_{kg}$ is defined as
\begin{equation}
  s_{kg}=\begin{cases}
    1&F_{kg}^0=F_{kg}^1=0\\
    \frac{\min\left( F_{kg}^0,F_{kg}^1 \right)}{\max\left( F_{kg}^0,F_{kg}^1 \right)}&\text{otherwise}.
    \end{cases}\label{eqn:fair-score}
\end{equation}

\subsection{Performance}\label{sec:perf}

To compute $F_{kg}^1$ and $F_{kg}^0$ in \Cref{eqn:fair-score}, we use the soft
F\textsubscript{1} score~\citep{kurniawan2025}. The score for class $k$ is
defined as
\begin{equation}
  2\frac{\sum_{i}\min\left( P_{ik},Q_{ik} \right)}{\sum_{i}\left( P_{ik}+Q_{ik} \right)}\label{eqn:sf1}
\end{equation}
where $P_{ik}$ and $Q_{ik}$ denote the ground-truth and the predicted
probability of class $k$ for instance $i$ respectively.

To evaluate overall performance with soft ground truths, we use the soft micro
F\textsubscript{1} score~\citep{kurniawan2025}. It is defined formally as
\begin{equation*}
  2\frac{\sum_{ik}\min\left( P_{ik},Q_{ik} \right)}{\sum_{ik}\left( P_{ik}+Q_{ik} \right)}.
\end{equation*}
This equation is similar to \Cref{eqn:sf1}, but the summations are performed
over both instances and classes.

\section{Difference in Fairness Interpretations}\label{sec:diff-fair-int}

Due to the different group definitions for both datasets, the interpretations of
fairness are also different. Consider an arbitrary group in SBIC. Here, a fair
model is one that performs similarly in identifying offensiveness on social
media posts that target and do not target the minority category corresponding to
the group. This is sensible because we do not want a model to identify offensive
posts less accurately when the posts target minorities. In contrast, for an
arbitrary group in TAG, a fair model is one that achieves similar levels of
performance for help-seekers that belong and do not belong to the group. This is
also sensible because we do not want a model to more frequently fail at
recognising relevant legal areas for a particular cohort~(e.g., seniors).
Thus, while there is a difference, the fairness definitions are still reasonable
for both cases.

\section{Configuration Sampling}\label{sec:conf-samp}

To get the results in \Cref{sec:mult-weight}, we randomly sample ten thousand
configurations for each of group- and class-wise aggregation. Each configuration
consists of a set of item weights and a scalar exponent $p$ that parameterises
the weighted generalised mean used for aggregation. Specifically, the weights
and exponent $p$ are drawn from a flat Dirichlet distribution and $U(-15,15)$
respectively. Smaller and larger values of $p$ make the generalised mean focus
more on lower and higher fairness scores respectively. In particular, as
$p\to\infty$ and $p\to-\infty$, generalised mean reduces to $\max$ and $\min$
respectively. In other words, the exponent $p$ controls how much the
lowest-scoring groups or classes are prioritised.

\section{HLV Training Methods}\label{sec:hlv-train}

We experiment with four HLV training methods:
\begin{enumerate}
  \item \methname{ReL}~(repeated labelling) which treats each annotation of an
    instance as a separate instance--class pair. In other words, an instance may
    appear multiple times in the training data, each with a different class.
    The loss function is standard cross-entropy.
  \item \methname{SL}~(soft labelling) which computes a class distribution for
    each instance and uses the distributions as soft ground truths for training.
    Each class probability is proportional to the number of annotators that
    select the class. Standard cross-entropy is used as the loss function.
  \item \methname{JSD} which computes class distributions similarly to
    \methname{SL} but uses Jensen-Shannon divergence as the loss function.
  \item \methname{SmF1} which also computes class distributions as \methname{SL}
    does but maximises the soft micro F\textsubscript{1} score during
    training~(see \Cref{sec:perf}).
\end{enumerate}
We refer the readers to the work by \citet{kurniawan2025} for more details on
these methods.

\section{Training Details}\label{sec:trn-details}

We implement all methods using FlairNLP~\citep{akbik2019}.\footnote{\url{https://flairnlp.github.io}} For RoBERTa, we tune
the learning rate and the batch size using random search, optimising only for
overall performance~(without fairness). \Cref{tbl:best-hparams} reports the best
values found. For OLMo, we use FlairNLP's default
hyperparameters\footnote{Learning rate and batch size are \num{5e-5} and 32
  respectively.} for computational reasons. We use LoRA~\citep{hu2022} to
finetune OLMo efficiently. All models are finetuned for 10 epochs. On a single
NVIDIA A100 GPU with 80GB of memory, the finetuning runs for:
\begin{itemize}
  \item less than an hour with RoBERTa for all methods except \methname{ReL} on
    SBIC and TAG,
  \item slightly over 2 hours with RoBERTa for \methname{ReL} on SBIC and TAG,
  \item slightly over 10 hours with OLMo for all methods except
    \methname{ReL} on SBIC,
  \item about 1.3 days with OLMo for \methname{ReL} on SBIC,
  \item about 9 hours with OLMo for all methods except
    \methname{ReL} on TAG, and
  \item about 2 days with OLMo for \methname{ReL} on TAG.
\end{itemize}
We truncate inputs longer than 512 tokens in TAG to reduce memory consumption,
affecting only 4\% of instances.

\begin{table}\small
  \centering
  \begin{tabular}{@{}llS[table-format=1.1e1]r@{}}
  \toprule
  Dataset                  & Method          & {Learning rate} & Batch size \\
  \midrule
  \multirow[c]{5}{*}{SBIC} & \methname{MV}   & 5.4e-07         & 4          \\
                           & \methname{ReL}  & 5.4e-07         & 4          \\
                           & \methname{SL}   & 1.7e-05         & 4          \\
                           & \methname{JSD}  & 4.6e-05         & 512        \\
                           & \methname{SmF1} & 4.6e-05         & 512        \\
  \midrule
  \multirow[c]{5}{*}{TAG}  & \methname{MV}   & 5.2e-06         & 16         \\
                           & \methname{ReL}  & 3.1e-05         & 8          \\
                           & \methname{SL}   & 3.1e-05         & 8          \\
                           & \methname{JSD}  & 6.1e-05         & 16         \\
                           & \methname{SmF1} & 2.6e-05         & 64         \\
  \bottomrule
  \end{tabular}
  \caption{Best hyperparameter values for RoBERTa.}\label{tbl:best-hparams}
\end{table}

\section{Effect Sizes}\label{sec:cies}

To complement our main result in \Cref{tbl:results}, we also report the effect
sizes between each HLV training method and \methname{MV} as a measure of
\emph{practical} significance~(as opposed to statistical). We report the
\qty{95}{\percent} bootstrap confidence intervals of Cohen's $d$ for both
performance and fairness evaluations in \Cref{tbl:cies}. The table shows that
all statistically significant results in \Cref{tbl:results} have an effect size
greater than \num{2.0}, which is considered ``huge''~\citep{sawilowsky2009}.
This observation demonstrates the practical significance of our findings.

\begin{figure}[t]
  \centering
  \includegraphics[width=0.95\columnwidth]{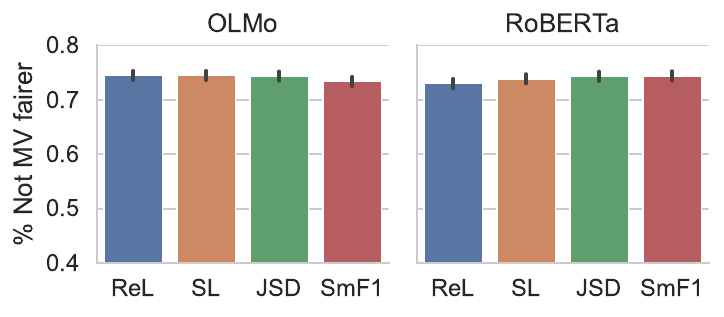}
  \caption{Fraction of randomly sampled configurations where \methname{MV} is
    not significantly fairer than each HLV training method on TAG. Whiskers
    indicate \qty{95}{\percent} bootstrap confidence
    intervals.}\label{fig:fair-conf-anal-TAG}
\end{figure}

\begin{figure}[t]
  \centering
  \includegraphics[width=0.95\columnwidth]{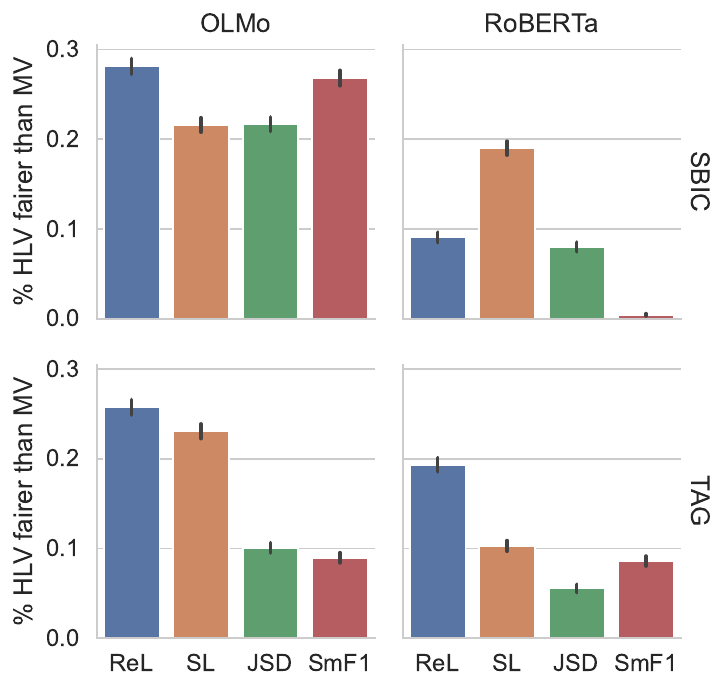}
  \caption{Fraction of randomly sampled configurations where each HLV training
    method is significantly fairer than \methname{MV}. Whiskers
    indicate \qty{95}{\percent} bootstrap confidence
    intervals.}\label{fig:fair-conf-anal-hlv-fairer}
\end{figure}

\begin{table*}\small
  \centering
  \sisetup{table-format=3.1}
  \begin{tabular}{@{}ll*{8}{S}@{}}
  \toprule
  \multirow[c]{4}{*}{Model} & \multirow[c]{4}{*}{Method} & \multicolumn{4}{c}{SBIC} & \multicolumn{4}{c}{TAG} \\
  \cmidrule(lr){3-6} \cmidrule(l){7-10}
  {} & {} & \multicolumn{2}{c}{Perf} & \multicolumn{2}{c}{Fair} & \multicolumn{2}{c}{Perf} & \multicolumn{2}{c}{Fair} \\
  \cmidrule(lr){3-4} \cmidrule(lr){5-6} \cmidrule(lr){7-8} \cmidrule(l){9-10}
  {}                          & {}   & {low} & {high} & {low}  & {high} & {low} & {high} & {low} & {high} \\
  \midrule
  \multirow[c]{4}{*}{OLMo}    & ReL  & 23.7  & 193.8  & -7.1   & 6.5    & 24.0  & 104.1  & 6.8   & 87.6   \\
                              & SL   & 29.4  & 239.9  & -8.1   & 3.1    & 16.7  & 80.5   & 5.9   & 61.8   \\
                              & JSD  & 26.1  & 180.6  & -7.9   & 0.5    & 5.4   & 44.3   & 3.6   & 23.7   \\
                              & SmF1 & 22.4  & 234.2  & -107.2 & -1.7   & 0.0   & 6.5    & 0.7   & 9.8    \\
  \midrule
  \multirow[c]{4}{*}{RoBERTa} & ReL  & 4.9   & 27.1   & -7.0   & 0.0    & 40.3  & 1957.4 & 18.8  & 486.1  \\
                              & SL   & -0.1  & 359.1  & 2.0    & 53.3   & 25.4  & 3154.8 & 9.5   & 58.6   \\
                              & JSD  & 1.1   & 12.0   & -2.5   & 3.1    & 38.4  & 939.3  & 10.7  & 46.7   \\
                              & SmF1 & 6.0   & 208.5  & -12.6  & -2.7   & 3.4   & 36.2   & 2.3   & 35.5   \\
  \bottomrule
  \end{tabular}
  \caption{\qty{95}{\percent} bootstrap confidence intervals of Cohen's $d$
    between each HLV training method and \methname{MV}. Values greater than
    \num{2.0} are considered ``huge'' effect
    size~\citep{sawilowsky2009}.}\label{tbl:cies}
\end{table*}

\section{More Multiple Fairness Configurations
  Results}\label{sec:mult-weight-more}

\Cref{fig:fair-conf-anal-TAG} shows the fraction of configurations where
\methname{MV} is not fairer than each HLV training method on TAG. The figure
shows a similar trend to that of SBIC reported in \Cref{sec:mult-weight}.

\begin{figure}[t]
  \centering
  \includegraphics[width=0.95\columnwidth]{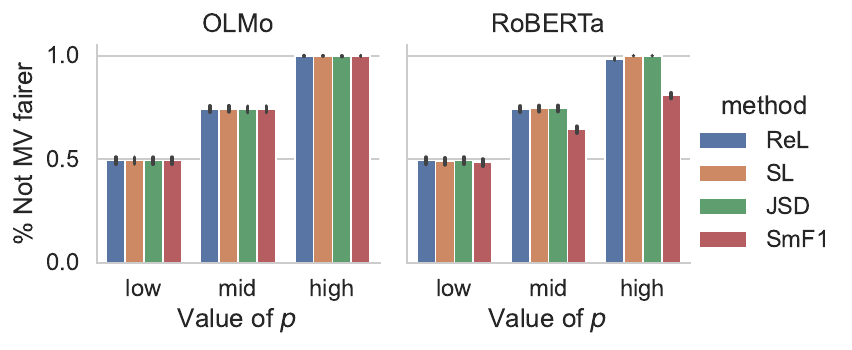}
  \caption{Fraction of randomly sampled configurations where \methname{MV} is
    not significantly fairer than each HLV training method on SBIC for various
    levels of exponent $p$ in the class-wise aggregation. Whiskers indicate
    \qty{95}{\percent} bootstrap confidence
    intervals.}\label{fig:fair-conf-anal-p-cls-sbic}
\end{figure}

\Cref{fig:fair-conf-anal-hlv-fairer} shows the fraction of configurations
where each HLV training method is fairer than \methname{MV}. We observe that
except for \methname{SmF1} with RoBERTa on SBIC, this fraction is substantial
for all dataset--model--method tuples. In particular, the fraction is over
\qty{25}{\percent} for \methname{ReL} with OLMo across the two datasets.

\begin{figure}[t]
  \centering
  \begin{subfigure}[b]{0.95\columnwidth}
    \centering
    \includegraphics[width=\textwidth]{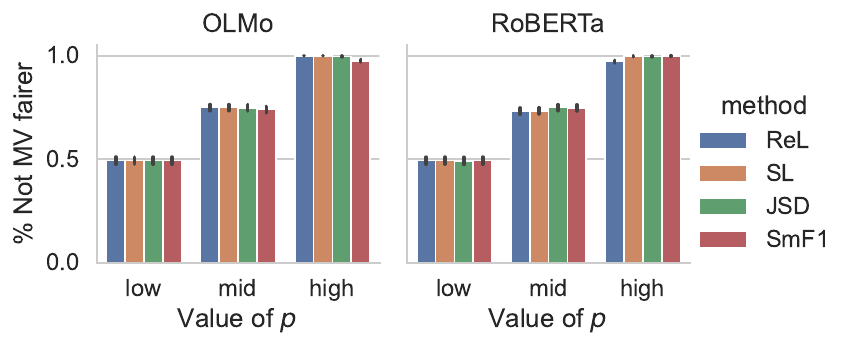}
    \caption{Group-wise}
  \end{subfigure}

  \vspace{3mm}

  \begin{subfigure}[b]{0.95\columnwidth}
    \centering
    \includegraphics[width=\textwidth]{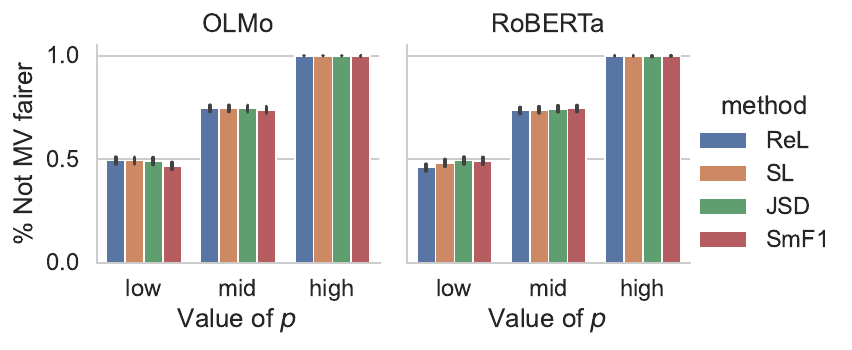}
    \caption{Class-wise}
  \end{subfigure}
  \caption{Fraction of randomly sampled configurations where \methname{MV} is
    not significantly fairer than each HLV training method on TAG for various
    levels of exponent $p$ in the group-~(top) or class-wise~(bottom)
    aggregation. Whiskers indicate \qty{95}{\percent} bootstrap confidence
    intervals.}\label{fig:fair-conf-anal-p-TAG}
\end{figure}

\section{More Fairness Configurations Analysis
  Results}\label{sec:fair-conf-anal-more}

This section shows the fraction of configurations where \methname{MV} is not
fairer than each HLV training method for various levels of exponent $p$.
\Cref{fig:fair-conf-anal-p-cls-sbic} shows the results on SBIC for the exponent
$p$ in the class-wise aggregation~(the group-wise counterpart is shown in
\Cref{fig:fair-conf-anal-p-grp-sbic}). In contrast,
\Cref{fig:fair-conf-anal-p-TAG} shows the same results on TAG for the exponent
$p$ in both the group- and class-wise aggregation. The figures show that the
trend is similar to that reported in \Cref{sec:fair-conf-anal}.

\section{Possible Explanation on Minority Annotations Improving Fairness on
  TAG}\label{sec:expl-minor-anns-TAG}

We offer a possible explanation for the finding in \Cref{sec:minor-anns} that the
minority annotations on TAG contain useful signals that improve fairness. TAG
was annotated by lawyers, each with a specific set of legal areas of
specialisation. Intuitively, some legal specialisations are more common than
others. We hypothesise that minority annotations are correlated with rare
specialisations, which also correlate with certain cohorts. This would imply
that incorporating the minority annotations would result in fairer models for
the defined cohorts.

\end{document}